\documentclass[acmtog]{acmart}

\AtBeginDocument{%
  \providecommand\BibTeX{{%
    \normalfont B\kern-0.5em{\scshape i\kern-0.25em b}\kern-0.8em\TeX}}}

\copyrightyear{2023}
\acmYear{2023}
\setcopyright{acmlicensed}\acmConference[AIES '23]{AAAI/ACM Conference on AI, Ethics, and Society}{August 8--10, 2023}{Montréal, QC, Canada}
\acmBooktitle{AAAI/ACM Conference on AI, Ethics, and Society (AIES '23), August 8--10, 2023, Montréal, QC, Canada}
\acmPrice{15.00}
\acmDOI{10.1145/3600211.3604690}
\acmISBN{979-8-4007-0231-0/23/08}

\usepackage{microtype}
\usepackage{graphicx}
\usepackage{subfigure}
\usepackage{booktabs} %

\usepackage{hyperref}
\usepackage{algorithmic}

\usepackage{amsmath}
\usepackage{mathtools}
\usepackage{amsthm}
\usepackage{xcolor}
\usepackage{float}
\newfloat{algorithm}{t}{lop}

\newcommand{\com}[1]{{\color{black!35}// #1}}
\newcommand{\methodacc}{MLAC}
\newcommand{\methodfullname}{Meta-Learned Adversarial Censoring}
\DeclareMathOperator*{\argmax}{\arg\!\max}

\usepackage[capitalize,noabbrev]{cleveref}

\theoremstyle{plain}

\theoremstyle{definition}

\theoremstyle{remark}

\usepackage[textsize=tiny]{todonotes}

\begin{document}

\title{Self-Destructing Models: Increasing the Costs of Harmful Dual Uses of Foundation Models}

\author{Peter Henderson}
\authornote{Both authors contributed equally to this research.}
\affiliation{%
  \institution{Stanford University}
  \city{Stanford}
  \country{USA}
}

\author{Eric Mitchell}
\authornotemark[1]
\affiliation{%
  \institution{Stanford University}
  \city{Stanford}
  \country{USA}
}

\author{Christopher D. Manning}
\affiliation{%
  \institution{Stanford University}
  \city{Stanford}
  \country{USA}
}

\author{Dan Jurafsky}
\affiliation{%
  \institution{Stanford University}
  \city{Stanford}
  \country{USA}
}

\author{Chelsea Finn}
\affiliation{%
  \institution{Stanford University}
  \city{Stanford}
  \country{USA}
}

\renewcommand{\shortauthors}{Henderson and Mitchell, et al.}

\begin{abstract}
A growing ecosystem of large, open-source foundation models has reduced the labeled data and technical expertise necessary to apply machine learning to many new problems.
Yet foundation models pose a clear dual-use risk, indiscriminately reducing the costs of building both harmful and beneficial machine learning systems. 
Policy tools such as restricted model access and export controls are the primary methods currently used to mitigate such dual-use risks.
In this work, we review potential safe-release strategies and argue that both policymakers and AI researchers would benefit from fundamentally new technologies enabling more precise control over the downstream usage of open-source foundation models.
We propose one such approach: the \emph{task blocking} paradigm, in which foundation models are trained with an additional mechanism to impede adaptation to harmful tasks without sacrificing performance on desirable tasks. We call the resulting models \emph{self-destructing models}, inspired by mechanisms that prevent adversaries from using tools for harmful purposes.
We present an algorithm for training self-destructing models leveraging techniques from meta-learning and adversarial learning, which we call \textit{meta-learned adversarial censoring (MLAC)}.
In a small-scale experiment, we show MLAC can largely prevent a BERT-style model from being re-purposed to perform gender identification without harming the model's ability to perform profession classification.
\end{abstract}

\maketitle

\section{Introduction}

A defining capability of large pretrained models (hereafter foundation models; FMs) is their ability to adapt to many downstream tasks in a few-shot manner---potentially improving performance and efficiency in domains with little training data~\citep{bommasani2021opportunities}.
Today, anyone with an internet connection can download a foundation model and adapt it to socially beneficial use-cases, like building better educational tools or improving access to justice.
However, a malicious actor can also adapt a foundation model to nearly any harmful use-case they desire.
For example, an oppressive government can take a powerful pretrained language model and adapt it to identify dissidents;
a rogue actor can adapt a pretrained object recognition system such that commercially available drones act as targeted loitering munitions; 
or a pretrained drug discovery system can be used for creating chemical or biological weapons, like neurotoxins~\citep{urbina2022dual}.
Unfortunately, due to their general-purpose nature, preventing such dual uses of foundation models is difficult.
This creates a tension between making these models widely available and ensuring that they are used in a safe and responsible way.

Currently, there are several approaches to mitigating the dual uses of FMs which can be divided into \textit{structural} safety mechanisms and \textit{technical} safety mechanisms. 
Structural mechanisms use licenses or access restrictions to prevent harmful uses; there is a broad spectrum of such structural release mechanisms.
Some have suggested a review board for selecting the structural release mechanism ~\citep{liang2022community-norms} while others have argued that open source access to foundation models is essential for safety research~\citep{black2022gpt}.
While structural release approaches aim to prevent malicious users from acquiring foundation models or providing legal remedies if they exceed the terms of their access, \textit{technical} strategies ensure that the model cannot be used for harmful purposes even if a malicious user is able to gain access to the model itself.
Current technical strategies aim to tune the model so that it is less likely to produce harmful content at inference time~\citep{bai2022training}, but do not consider the case where adversaries have access to model parameters.

In this work, we review these strategies, noting that no strategy on its own is able to prevent harmful dual uses of FMs.
In particular we note the disconnect between the goal of many structural safety mechanisms and the new reality of open-source foundation models: structural safety strategies aim to prevent a malicious actor from gaining access to the model parameters altogether.
In recent months, however, powerful open-source models have been released to the public, including Meta's Llama model which was leaked online despite a restricted access policy~\citep{touvron2023llama,vincent2023llama}.
Such developments demand changes to the threat model of malicious FM usage, specifically, that eventually model parameters will become generally accessible.
Unlike the assumptions of current safety strategies, there should then be a last layer of defense that renders the model itself as harmless as possible.
We argue that we need more \textit{technical} strategies to supplement \textit{structural} strategies to reduce the ability for adversaries to use and adapt foundation models for harmful tasks: even when they have access to model parameters.
Where existing access restrictions must navigate the tension between openness and safety, we seek to provide a new research pathway for reducing (and in some cases obviating) this tension.

We suggest one such new path forward: \textit{self-destructing models}.
Self-destructing models are trained via a task blocking method that impedes the adaptation of the model to a harmful task without impairing the model's ability to be used for its original intended purpose.
By increasing the compute, data, and talent required to adapt public models to harmful tasks, self-destructing models have the potential to supplement access controls and other safety mechanisms.
We demonstrate a task-blocking mechanism using meta-learning for training a self-destructing model. We find that meta-learning is an essential step in reducing an adversary's ability to tune a model for a harmful task. Simple adversarial losses~\citep{edwards2015censoring}, often used in current technical strategies,
do not significantly reduce the costs of harmful adaptation. 
We hope that the proposed mechanism forms an initial step toward developing new safe release strategies even under the assumption that model parameters become available to adversaries.

Below, we first review the state of current safe-release approaches and their shortfalls, making the case for a shift in the threat model to make model parameters as harmless as possible even with model access.
Second, we define the \textit{task blocking} problem and evaluation metrics as well as \textit{self-destructing models}.
Third, we describe an initial algorithm, \methodfullname{} (\methodacc{}), for training self-destructing models, evaluating its ability to impede fine-tuning a language model to perform demographic information extraction. Fourth, we identify key directions for future research in the development of self-destructing models.

\section{Reviewing the Risks and Mitigation Strategies for Dual Uses}

Foundation models can be and, importantly, \textit{have been} used for harmful purposes unforeseen by their creators in recent years. They have been fine-tuned on hate speech and deployed to 4chan~\citep{vincent2022youtuber}; hackers have released methods to bypass ChatGPT's safety filters so that it can be used to help generate malware and spam~\citep{hackersgpt}; stable diffusion models have been fine-tuned to generate abusive imagery~\citep{aideepfake}.

Researchers, practitioners, and policymakers are increasingly searching for new ways to prevent machine learning models from being used for these harmful dual purposes---e.g., \citet{solaiman2023gradient}, \citet{brundage2020toward}, \citet{whittlestone2019tension}, \citet{shevlane2022structured}, \citet{brundage2018malicious}, and many others. 
Proposed tools have included export controls, controlled or restricted release strategies, using terms of service or licensing, and alignment and fine-tuning for safety.
In this section, we briefly examine each of these methods and discuss potential gaps in relying on each strategy. We consider both \textit{structural} methods (e.g., export controls, use of licensing, and access restrictions), and \textit{technical} methods (e.g., alignment fine-tuning).

\begin{table*}[!htbp]
    \centering
    \begin{tabular}{c|c|c}
    \toprule
        Approach & Examples & Challenges\\
        \midrule
        Export Controls & United States Export Controls on AI hardware & Imprecise, reduced hardware costs, open-source models\\
        Controlled Release & API-only access, Release by request/agreement & Open-source models, leaks, monitoring difficulties\\
        Licensing & OpenRAIL & Requires monitoring and enforcement action, leaks\\
        Filtering, Alignment & Reinforcement Learning from Human Feedback & Can be bypassed by fine-tuning or prompt engineering\\
        \bottomrule
    \end{tabular}
    \caption{A review of current or proposed approaches to safe foundation model release.}
    \label{tab:my_label}
\end{table*}

\subsection{Structural Methods}

\noindent \textbf{Export Controls.} Recently, researchers, such as \citet{flynn2020recommendations}, have recommended that the United States consider export controls on hardware related to AI, including NVIDIA A100 GPUs, to restrict certain actors' capacity to train powerful AI models that require substantial computational resources. In 2022, the United States imposed these export controls on AI-related hardware and hardware-manufacturing equipment, following researchers' suggestions~\citep{federalregisterexportcontrols}. 

Such export controls may help restrict pre-training of foundation models---a use case which requires large amounts of specialized hardware, but they do not necessarily restrict inference-time computing and small-scale adaptation once model parameters are available. Even the largest foundation models can now be deployed or adapted on commodity hardware using techniques such as adapters \citep{hu2021lora}, 8-bit \citep{dettmers2022llmint8}, and even 4-bit \citep{dettmers2022case} quantization, and other optimization strategies. A recent open-source project was able to run multi-billion parameter LLaMa models on a MacBook Pro with near-equal performance to some state-of-the-art closed-source models, using these techniques.\footnote{\url{https://github.com/ggerganov/llama.cpp}}
As a result, hardware export controls may no longer be sufficient to prevent the efficient adaptation of foundation models or the large-scale deployment of pre-trained models, nor can they prevent malicious actors located in countries not included in the export control regime.

The U.S. government has also put in place export controls on certain \textit{software} and \textit{models} with specific harmful dual uses.
For example, in a 2020 rulemaking, the Department of Commerce Bureau of Industry and Security (BIS) restricted export of software that can be used for automated geospatial analysis. Under this regulation the model is controlled if it meets four criteria: (1) it provides a graphical user interface to identify objects in geospatial imagery; (2) it ``reduces pixel variation by performing scale, color, and rotational normalization on the positive samples''; (3) it ``[t]rains a Deep Convolutional Neural Network to detect the object of interest from the positive and negative samples''; (4) it ``[i]dentifies objects in geospatial imagery using the trained Deep Convolutional Neural Network by matching the rotational pattern from the positive samples with the rotational pattern of objects in the geospatial imagery.''
But such highly specific export controls do not cover general-purpose foundation models (and associated training software). 
In fact, a recent demonstration showed how to adapt a CLIP model~\citep{radford2021learning} exactly for analyzing satellite imagery in an easy way using all open-source software~\citep{cliprisd}.
\citet{flynn2020recommendations} argued that applying export controls to general-purpose foundation models would be ineffective due to the ease of violating export controls through the same mechanisms as software piracy, as well as the harmful impacts to innovation that such restrictions could have.

Overall, while export controls may be effective in restricting access to large-scale chipsets or certain software, once adversaries can gain access to open-source (or leaked) foundation model parameters they can be readily adapted to harmful dual-uses.

\medskip

\noindent \textbf{Access Control.} Controlled release or restricted access strategies are another set of structural mechanisms that can supplement export controls and reduce malicious actors' ability to access models \citep{shevlane2022structured,solaiman2019release,ovadya2019reducing}.

One such approach is to make the model accessible only by agreement. This involves vetting potential users and requiring them to sign a restrictive terms of service before gaining access to the model. For instance, Meta's OPT-175B model and Llama both employ this approach~\citep{zhang2022opt,zhang2022democratizing,touvron2023llama}. This access restriction approach is attractive as it does not require hosting any centralized infrastructure for serving model queries. It only requires one-time vetting of the users requesting model access. However, as evidenced by the recent Llama model leak onto BitTorrent~\citep{vincent2023llama} and HuggingFace,\footnote{\url{https://twitter.com/ClementDelangue/status/1632948540245671936}} this approach is susceptible to unauthorized dissemination, effectively negating access control efforts.

Another approach is to never release the model at all, but provide access via an application programming interface (API). Many companies, such as OpenAI, Anthropic, Cohere, and AI21 adopt this approach to protect their trade secrets and prevent harmful dual uses. This approach prevents direct access to model weights, preventing uncontrolled dissemination and retaining the ability to cut off access to malicious users at any time.
However, this approach requires monitoring of API usage to detect and revoke access when abused, as well as considerable resources to maintain.
Providing such an API may not be possible for researchers and entities without access to centralized model-hosting infrastructure. 

Additionally, as open-source efforts continue to match the performance of these closed-source models, the effectiveness of any access control approaches may decrease.
Access control approaches require all model creators capable of training similarly capable foundation models to be in agreement on the mechanism for release. 
If one equally-capable foundation model is available as open-source, malicious actors can simply turn to this alternative. 

\medskip

\noindent \textbf{Terms of Service/Sale (ToS) and Licenses.} Closely tied to access controls are licensing agreements to prevent harmful dual-uses. These agreements place restrictions on who can use the model, for what purpose, and in what format. For example, OpenRAIL~\citep{openrail2022} and similar licenses impose several usage limitations to prevent users from using the model for defamation, spreading disinformation, providing medical advice, or for use by law enforcement. Such terms of service (ToS)-based approaches are also used in other settings, such as by Boston Dynamics, which prohibits modifying its robots for lethal capability and reserves the right to prevent any misuse.\footnote{\url{https://twitter.com/BostondDynamics2021/status/1362921918781943816}}

However, relying solely on licensing agreements assumes that malicious actors would respect them and that legal action against violators is possible. Unfortunately, this approach faces several challenges. Firstly, harmful actors may be located in countries that do not enforce licensing requirements. Further, it may be challenging to identify malicious actors and issue a cease-and-desist request. Finally, model creators may not have the resources to monitor and enforce compliance with licensing agreements.

Overall, licensing agreements face the same challenges as other structural restrictions. They require the resources, and international reach, for enforcement.

\subsection{Technical Strategies}

Unlike structural strategies, we classify \textit{technical strategies} as those that modify foundation models directly to make it more difficult to use them for harmful purposes. Existing technical strategies focus on tuning models to prevent them from outputting harmful content at inference time or adding content filters to block potentially harmful outputs.

\medskip

\noindent \textbf{Safety Filters.} Some models come with safety filters that scan model outputs for harmful content and then redact the output. Stable Diffusion models use this approach to replace offensive content generated by the model with a blank image by default~\citep{rombach2021highresolution}.
However, for open-source models safety filters can simply be removed by deleting a few lines of code. This has led users on Reddit to post tutorials like ``How to remove [Stable Diffusion's] safety filter in 5 seconds.''\footnote{\url{https://www.reddit.com/r/StableDiffusion/comments/wv2nw0/tutorial_how_to_remove_the_safety_filter_in_5/}} Other researchers have noted that the filter itself is easily bypassed even without access to directly modify the code~\citep{rando2022red}. While safety filters can be effective and integral parts of a safe model release, they are more effective when coupled with other structural mechanisms like restricted or API-only model access.

\medskip

\noindent \textbf{Safety Tuning and Alignment.} Alternative approaches such as reinforcement learning from human feedback tune the model itself to be less harmful~\citep{bai2022training}. Sometimes these approaches fall into a larger class of methods under the moniker \textit{AI alignment}. Since these methods directly train the model to be more difficult to use for harmful purposes at inference time, they are an essential part of a safe release strategy---either for open-source models or for models coupled with a structural release restriction. Though they make the model parameters more difficult to use for harmful tasks, they can be bypassed in two ways.

First, prompt engineering can be used to put models in a state that nonetheless allows them to be used for harmful purposes.
For example, hackers now sell prompts and methods to bypass alignment processes and filters for OpenAI's series of models~\citep{hackersgpt}. This allows would-be malicious actors to generate phishing emails and malware with the model, despite its use restrictions.

Second, open-source models can be fine-tuned to remove these restrictions. In one such instance, the open-source GPT-J model was fine-tuned on 4chan data (mainly consisting of toxic content and hate speech) and deployed to post to the forum~\citep{vincent2022youtuber}.

In the remainder of this work, we describe and evaluate an approach to mitigating this second method of bypassing existing technical model protections.

\subsection{The Need For New Technical Mitigation Strategies}

The strategies discussed above are individually imperfect; however, each contributes to increasing the costs of successfully co-opting foundation models for harmful dual uses. As access to increasingly capable models becomes commonplace—either through leaks or open-source releases—it is crucial to ensure that the underlying model parameters themselves are optimized for safety as a last line of defense. Structural barriers, such as access restrictions and terms of service, can become ineffective as model weights are distributed through services like BitTorrent.

As regulators recognize the potential dangers associated with increasingly capable systems, it is becoming evident that they will take action to address the risks. 
One E.U. AI Act proposal would see liability placed on open-source models, incentivizing restricted access approaches. 
Others argue that such a move would stifle innovation and make it more difficult to develop safer overall models~\citep{hacker2023regulating,engler2022,black2022gpt}.
As \citet{black2022gpt} write, ``open access to [FMs] is critical to advancing research in a wide range of areas—particularly in AI safety, mechanistic interpretability, and the study of how [FM] capabilities scale.''
Yet while more widely available FMs certainly enable greater accessibility, auditability, and understanding of these powerful models, making FMs widely available for downstream adaptation without restriction comes at some cost to safety.

Despite the benefits of open-source releases, if open-source models are regularly adapted for harmful purposes, the pendulum of regulation could swing toward the more restrictive regime as regulators look to available structural tools like access restrictions.
To supplement the policy options available to regulators, and to increase the safety of foundation models by default, we encourage more research to expand the toolbox of technical approaches to ensure that model parameters are as safe as possible, even when they are leaked or openly available.
We introduce a new class of methods for this toolbox: \textit{task blocking} for \textit{self-destructing models}. These methods are not perfect, but add another layer of protection when combined with other approaches.

\section{Task Blocking \& Self-Destructing Models}
\label{sec:taskblocking}

\begin{figure}
    \centering
    \includegraphics[width=.96\columnwidth]{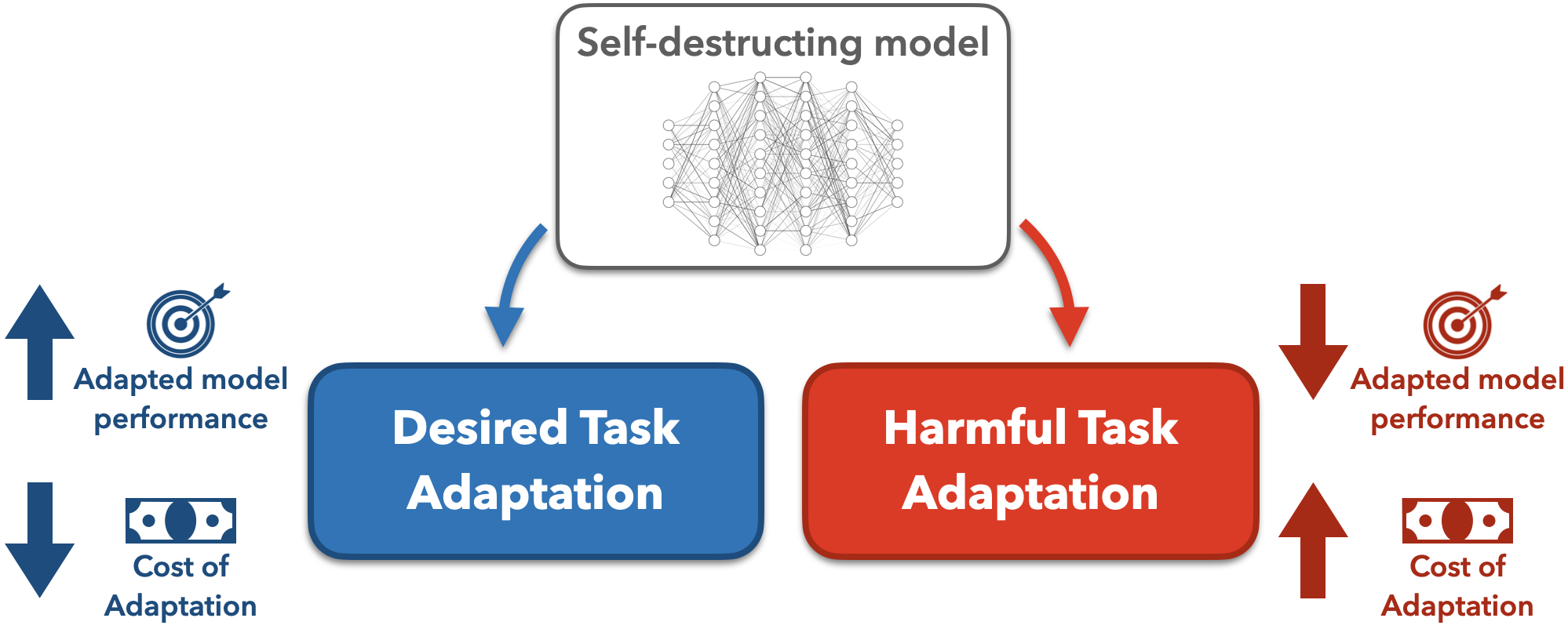}
    \caption{An ideal self-destructing model would boost performance and reduce adaptation costs relative to training from scratch only for desired tasks, while impeding learning of harmful tasks.}

        \label{fig:setting}
\end{figure}

The goal of task blocking is to create models that increase the costs of fine-tuning on harmful downstream tasks such that an adversary would rather start from scratch than use the pretrained model, while remaining useful for desired tasks (see Fig.~\ref{fig:setting}).
The resulting models are ``self-destructing models'' which impede adaptation on harmful dual-uses by increasing the costs of the harmful use. 
In this section, we more precisely define our problem setting and describe an initial algorithm for it.

\subsection{The Task Blocking Problem}
We assume that an adversary aims to adapt a pretrained model $\pi_\theta$ (where $\theta$ are model parameters of model $\pi$) to a harmful task, searching for the best adaptation procedure $f$ among a set of adaptation procedures $\mathcal{F}$ in order to find the one that maximizes harmful task performance. 
Adaptation procedures in $\mathcal{F}$ may include simple fine-tuning, a hyper-parameter search over fine-tuning procedures, as well as other more advanced adaptation mechanisms that we leave to future work.
The goal of task blocking is to produce a self-destructing model with parameters $\tilde\theta$, which performs similarly to a standard pre-trained model on a set of desired tasks while being more costly to successfully adapt to harmful tasks.\footnote{While the goal of a self-destructing model is to reduce performance on harmful tasks after fine-tuning, it should enable high quality \textit{predictions} or \textit{fine-tunability} for desired tasks. Our experiments explore the prediction goal, and we leave exploration of preserving fine-tunability for future work.}

We define two regimes to increase costs: (1) increase data costs by decreasing sample efficiency; (2) increase compute costs by slowing convergence of the training process. 

\medskip

\noindent \textbf{Data Costs.} In the first regime, we assume that the adversary has little data to adapt an FM to their harmful task and that the cost of gathering more data is high. 
A hallmark trait of traditional FMs is effective few-shot adaptation, learning rapidly from small, fixed-sized datasets. A self-destructing FM, on the other hand, should provide few-shot performance comparable to a randomly initialized model. We define the \textit{few-shot performance improvement} of an FM with parameters $\theta$ as the performance gain over a randomly initialized model, both with a fixed adaptation procedure search budget.
This can be represented as the following formula:

\begin{equation}
    \mathcal{E}_{data}^n(\theta) = \max_{f\in\mathcal{F}} \mathcal{M} \left(f(\theta, D_n)\right) - \max_{f\in\mathcal{F}}{\mathcal{M} \left( {f(\theta^r, D_n)}\right)}, 
    \label{eq:data}
\end{equation}
where $\mathcal{M}$ is the performance metric (where higher is better), $n$ is the number of data points available, $D_n$ is an adaptation dataset of $n$ examples from the task of interest, and $\theta^r$ is a randomly-initialized model. $f \in \mathcal{F}$ is an adaptation procedure drawn from a fixed distribution. The size of $\mathcal{F}$ loosely corresponds to the adversary's resource budget for adaptation. Note that the $\max$ in Equation~\ref{eq:data} encapsulates hyperparameter optimization over the adaptation distribution.
$\mathcal{E}_{data}=\frac{1}{N}\sum_n^N \mathcal{E}_{data}^n$ is the average sample-wise regret between the FM parameters $\theta$ and a random re-initialization $\theta^r$ after each follows the same adaptation procedure $f(\cdot)$ on a fixed-sized dataset $D_n$. An ideal self-destructing model has $\mathcal{E}_{data} \le 0$, meaning the model is no more data efficient than a randomly-initialized model for the (presumably harmful) task of interest.

\medskip

\noindent \textbf{Compute Costs.}
\label{app:compute}
If data is cheap or plentiful, it may be difficult to prevent an adversary from learning the task since perhaps even a random model can learn the task with the amount of data available.
In this data regime (large amount of cheap data), the benefit of an FM is improved compute efficiency, rather than increased accuracy. Here, we would define the FM's \textit{compute cost improvement @$p$} as the amount of compute saved by using the FM over a randomly initialized model to achieve performance $p$, where $p$ may measure accuracy, loss, or another metric and compute could be measured in FLOPs, train steps, hyperparameters searched, wall clock time, etc. While in the previous setting, we fix the \textit{dataset size} and blocking aims to reduce performance, in this setting, we fix the \textit{performance} and blocking aims to increase compute costs. The goal of task blocking in this case is to prevent any compute cost improvement over a random initialization when adapting the self-destructing model to a harmful task, while retaining compute cost improvement for desired tasks. Formally, compute cost improvement @$p$ is given as
\begin{equation}
    \mathcal{E}_{compute}^p(\theta) = \mathcal{C}(\mathcal{F}, \theta^r, p) - \mathcal{C}(\mathcal{F}, \hat{\theta}, p)
\end{equation}
where $\mathcal{C}$ measures the compute cost of applying adaptation procedures from family $\mathcal{F}$ to random parameters $\theta^r$ or FM parameters $\theta$ until a model with performance level $p$ is found.

However, for the purposes of this work, we focus on data costs, studying methods for reducing few-shot performance improvement for harmful tasks. We leave analysis of compute cost improvement reduction to future work.

\medskip

\noindent \textbf{Defining Harmful Dual Uses.} A large body of work has pointed to inherently harmful uses that FM creators may wish to block: from creating neurotoxins~\citep{urbina2022dual} to race detection~\citep{olson2022quiet}.
In our work we assume that a harmful dual use is \emph{known} and \emph{defined}.
That is, the self-destruct mechanism will have data to approximate the dual use and actively encode a mechanism to block it. This requirement inherently requires a normative definition of harmful dual uses. 
As in other threat modeling exercises and mechanisms for removing harmful content from models, model creators will have to identify the set of tasks to be blocked.
Creating self-destructing models may impede their use for harmful purposes counter to the model creator's values, but it is up to the model creator to determine those values.
While defining harmful tasks \emph{a priori} may be difficult, this work reflects a ``red teaming'' approach to harm prevention, common in security contexts. That is, model creators play the role of an adversary to identify and prevent harms.
This can function as a complement to other access control methods, providing more confidence that certain known harmful tasks are blocked.

\noindent \textbf{Relationship to Other Technical Safety Mechanisms.} Reinforcement learning from human feedback (RLHF), and other similar approaches, have been used to mitigate the harms that model can have at inference time~\citep{bai2022training}. While RLHF aims at ensuring that agents are as harmless as possible at inference time, the goal of self-destructing models and task blocking is to make it difficult to undo these safety mechanisms and co-opt the model even with access to model parameters and adaptation. These are complementary approaches and can be used concurrently to make the model parameters as safe as possible overall. Essentially, the aim is to maintain the model's harmlessness for as long as possible, even when an adversary has direct access.

\subsection{\methodfullname{}}
\label{sec:method}

\begin{algorithm}
    \caption{\methodacc{} Training Procedure}
    \label{alg:mlac}
    \begin{algorithmic}[1]
     \STATE \textbf{Input:} pretrained model $m = w_d \circ \pi_\theta$, desired task dataset $D_d$, harmful task dataset $D_h$, adaptation methods $\mathcal{\tilde F}$, adaptation steps $K$, learning rates $\eta$, $\eta_h$, $\eta_d$
        \STATE \textbf{Initialize:} Adversarial harmful task head $w_h$ and learning rate $\alpha_h$, with $\phi = \{w_h, \alpha_h\}$; initial blocked params ${\tilde\theta} \gets \theta$
        \FOR{$n$ steps}
            \STATE Sample adaptation procedure $\tilde f_k \sim \mathcal{\tilde F}$
            \STATE Sample data batches $b_d \sim D_d$, $\{b_{h}^k\} \sim D_h$, $b_{h} \sim D_{h}$
            \STATE $\{\theta_k\}, \{w_h^k\} \gets \tilde f_k(w_h \circ \pi_{\tilde\theta}, \{b_{h}^k\} , \alpha_h)$ \hfill \com{do inner loop}
            \STATE $\ell^{h}_k = \mathcal{L}_{h}(w_h^k \circ \pi_{\theta_k}, b_h)$, $\forall k$ \hfill \com{outer loop harmful NLLs}
            \STATE $\ell^{d} = \mathcal{L}_{d}(w_d \circ \pi_\theta, b_d)$ \hfill \com{desired NLLs}
            \STATE $\tilde\theta \gets \tilde\theta - \eta \nabla_\theta \left(\ell^d - \frac{1}{K}\sum_{k}^{K} \ell^h_k\right)$ \hfill \com{update blocked model}
            \STATE $\phi \gets \phi - \eta_h \frac{1}{K}\sum_{k=1}^K \nabla_\phi \ell^h_k$ \hfill \com{update adversarial params}
            \STATE $w_d \gets w_d - \eta_d \nabla_{w_d} \ell^d$  \hfill \com{update desired task head}
        \ENDFOR
    \end{algorithmic}
\end{algorithm}

To prevent successful adaptation of pretrained models to harmful tasks, we describe \textit{\methodfullname{} (\methodacc{})}, a meta-training procedure that aims to eliminate any useful information about the harmful task in the model's parameters \textit{even after fine-tuning on that task}. Given a desired task dataset $D_d$ and harmful task dataset $D_h$, \methodacc{} learns a feature extractor $\pi_{\tilde\theta}$ that is effective for the desired task but cannot be effectively used or efficiently fine-tuned to perform the harmful task.

In the \textit{inner loop} of each meta-training step, the feature extractor and an adversarially learned prediction head $w_h$ are adapted to the harmful task with several steps of gradient-based adaptation with an adversarially learned learning rate $\alpha_h$. The adaptation procedure $\tilde f$ used at each meta-training step is sampled from $\mathcal{\tilde F}$, a proxy for the true adversary's adaptation class $\mathcal{F}$. 
In this case, we narrow $\mathcal{\tilde F}$ to be different fine-tuning approaches with close-to-optimal hyperparameters (e.g., Adam for $K$ steps and learning rate $\alpha_h$). 
In the \textit{outer loop}, the adversarial parameters $\phi = \{w_h, \alpha_h\}$ are trained to minimize the harmful task negative log likelihoods of the adapted models, while the blocked parameter initialization $\tilde\theta$ are trained to maximize the harmful task negative log likelihoods of the adapted models. 
We also must counteract the self-destruct mechanism with something that will prevent unlearning of the entire network. In this work, we simply optimize for a given desirable task as the counter-balance by minimizing $\ell^d$, which updates both the desired task head $w_d$ and the representation parameters $\tilde\theta$.
See Algorithm~\ref{alg:mlac} for the complete training procedure. Note that in practice, we use Adam rather than SGD in the outer loop to optimize $\tilde\theta$, adversarial parameters $\phi$, and desired task output head $w_d$. We use \texttt{higher}~\citep{grefenstette2019generalized} for implementing the bi-level meta-learning process.

\medskip

\noindent \textbf{Calibration.} We also add another mechanism to strengthen the inner-loop adversary. In binary classification tasks, maximizing the loss of the harmful task may lead to a degenerate optimum where labels are flipped, which leaks information about the harmful task. 
To prevent this outcome, we also optimally calibrate the logits via a simple linear projection ($w$) solved via differentiable convex optimization~\citep{diamond2016cvxpy,cvxpylayers2019}. Thus at step $k$ of the inner loop we solve the maximum likelihood problem:
\begin{align}
\label{eq:calibration}
       w_c^k = \argmax_{W} \sum_i^{|b_h|} \left[ \text{logsoftmax}\left[\left(W\hspace{-0.25mm}\circ \hspace{-0.25mm} m^k\right) \hspace{-0.75mm}(x_i)\right]^\top \hspace{-1mm}y_i \right] \nonumber \\ \text{s.t.}  -1 \le W \le 1,
\end{align}
where $m^k = w_h^k \circ \pi_\theta^k$ is the blocked model after $k$ steps of adaptation using the adversarial harmful task head and learning rate.
Thus this projection updates line 7 of Alg.~\ref{alg:mlac} to $\ell^{h}_k = \mathcal{L}_{h}(w_c^k \circ w_h^k \circ \pi_{\theta_k}, b_h)$. We also refer to calibration as \textit{head adjustment}, as it essentially refines the linear function computed by the final output head.

\begin{figure}[t]
    \centering
    \includegraphics[width=.99\columnwidth]{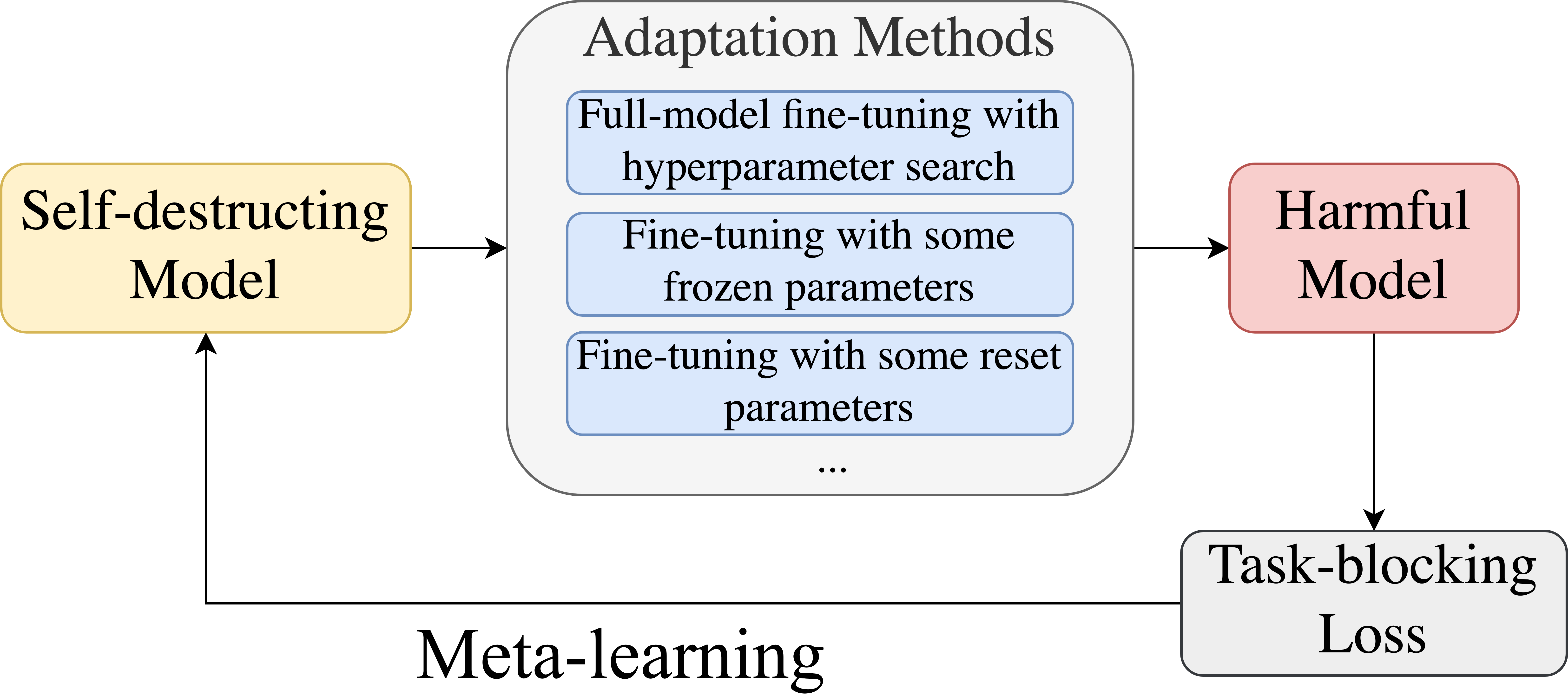}
    \caption{High-level visualization of the meta-learning process.}
    \label{fig:mlac_intuition_1}
\end{figure}

\noindent \textbf{High-level Intuition.} Figures~\ref{fig:mlac_intuition_1} provides a visualization of this method. At each step, the self-destructing model samples from possible adaptation methods that could be used to adapt the model to a harmful dual use. This multi-step loss is then inverted in a meta-learning step to prevent the model from being easily adapted in this sampled fashion.

\begin{figure}[t]
    \centering
    \includegraphics[width=.99\columnwidth]{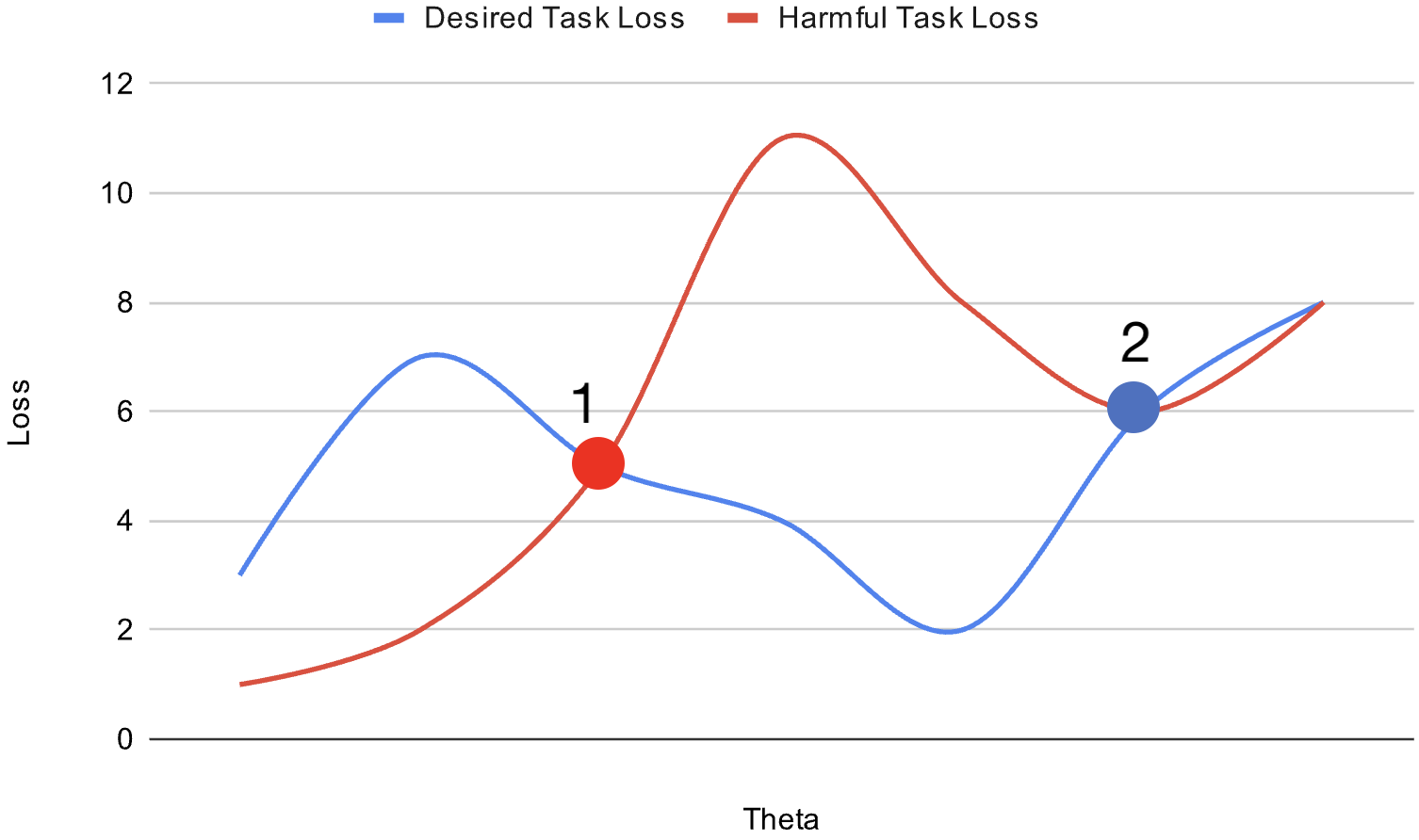}
    \caption{High-level optimization perspective of the MLAC procedure. A foundation model placed in point 1 would easily be tuned via gradient descent for both the harmful task loss and the desired task loss global optimum. On the other hand a foundation model in point 2 would easily reach the desired task optimum, but is more likely to be stuck in a local optimum for the harmful task.}
    \label{fig:optimization_perspective}
\end{figure}

From an optimization perspective, the goal is to identify a parameter space where adaptation to desired tasks is relatively simple via standard adaptation techniques, but the same part of the parameter space might be a low-utility local optimum or saddle-point that is more difficult to escape for the harmful task. This can be seen as a simplified visualization in Figure~\ref{fig:optimization_perspective}. Of course, adaptation methods can be created to reset parts of the network such the global harmful optimum can be recovered (in the extreme resetting most of the network to escape the local optimum). However, this will decrease the utility of the expensive pre-training and increase the costs to adversaries, adding another tool in the toolkit against harmful dual-uses.

\section{Experiments}

The goal of our experiments is to assess the ability of MLAC and several simple baselines to reduce the few-shot performance improvement of a pre-trained FM.\footnote{Code is available at \href{https://github.com/Breakend/SelfDestructingModels}{https://github.com/Breakend/SelfDestructingModels}.}

\subsection{Dataset.} In a demonstrative experiment, we utilize an existing dual-use dataset in the de-biasing literature, ``Bias in Bios''~\citep{de2019bias}. The dataset consists of professional biographies. Each biography has a label that is the gender identity of the biography's subject as well as the profession being discussed.
We split the data into a train, validation, and evaluation set.
We consider the ``desirable task'' for which we want to maintain good performance as the profession detection task. 
We wish to block the gender identification task. 
On the original dataset, we find that a random model can learn gender classification to over 90\% accuracy with only 10 examples, leaving only marginal ability for an FM to improve in data efficiency.
Thus, to make the FM more beneficial, we replace all pronouns with ``they/their,'' similar to the censored dataset in the original data.
While this task pair has traditionally been used for de-biasing, and while we mainly use this as an initial demonstration, there may be valid reasons for preventing an adversary from detecting demographic information from text.
In countries where anti-minority action is common, automated systems that identify demographics may cause serious harm.
For example, a country may wish to identify people of a certain religion, sexual orientation, or other identity group in automated and ultimately harmful ways. Or the state may wish to identify another feature that is highly correlated with identity which will lead to the same harms.\footnote{\href{https://www.brennancenter.org/sites/default/files/Technology\%20Experts\%20Letter\%20to\%20DHS\%20Opposing\%20the\%20Extreme\%20Vetting\%20Initiative\%20-\%2011.15.17.pdf}{Technology Experts Letter to DHS Opposing the Extreme Vetting Initiative, 2017.}} 

\subsection{Protocol.} For all experiments, we run 50k steps of \methodacc{} meta-training on the training set.
At test time, we take the resulting self-destructing model and run it through a rigorous hyperparameter search to maximize fine-tuning performance on the harmful task.
We allow hyperparameter searches with 50 fine-tuning trials, using the tree-structured Parzen Estimator \citep{bergstra2011algorithms} in the \texttt{hyperopt} software package~\citep{bergstra2013hyperopt}. We search over learning rate, batch size, maximum number of steps, and freezing of intermediate representation layers. 
For this process, we subsample the validation set to simulate an adversary with a dataset of size $N$. This subsampled validation set is used as the training set for the adversary. We then use the entire evaluation set to evaluate the adversary's performance on held-out data and for hyperparameter tuning.
We make the conservative assumption that the adversary can perform hyperparameter tuning using the \textit{population}, even if the amount of data for fine-tuning itself is limited. This choice weighs heavily in the adversary's favor, disadvantaging the self-destruct method.
We repeat the hyperparameter search process 6 times with different random seeds and data subsets. This yields confidence intervals over different adversaries training on different subsets of the data.

\begin{figure}[t]
    \centering
    \includegraphics[width=.87\columnwidth]{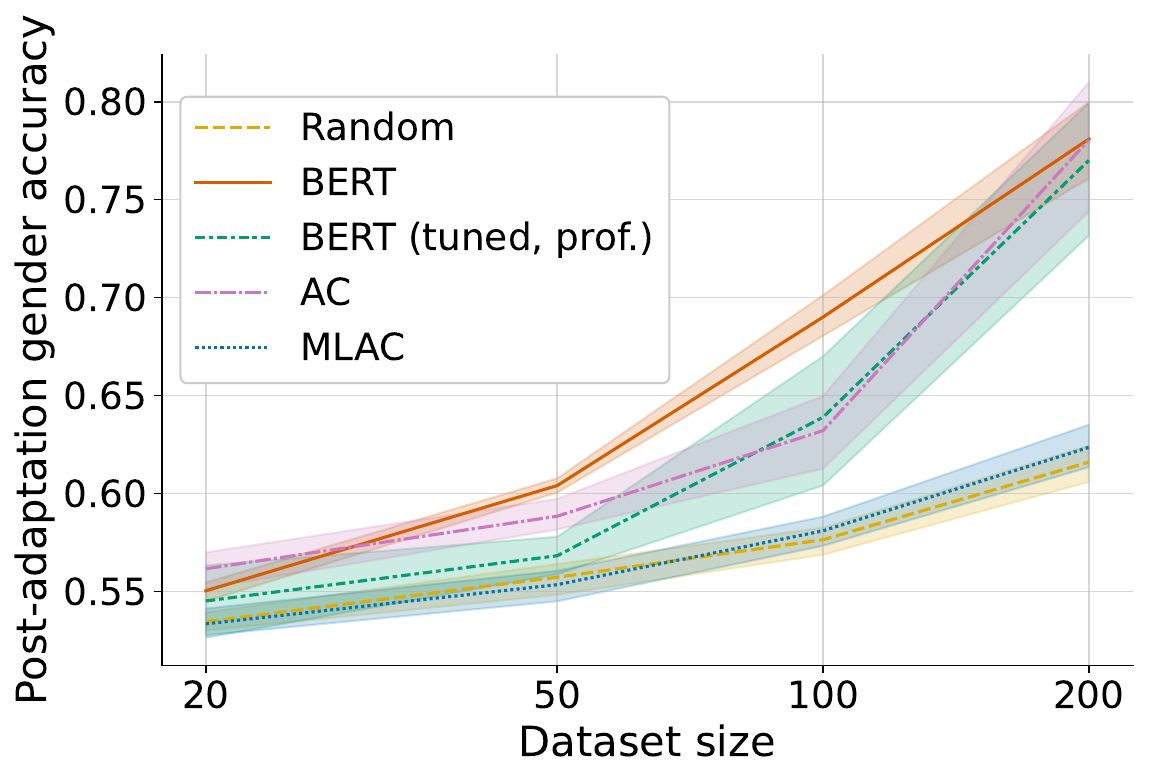}
    \caption{Harmful task (gender identification) performance after fine-tuning. \methodacc{} shows fine-tuning performance similar to a randomly-initialized model, while adversarial censoring (AC) \citep{edwards2015censoring} does not prevent effective fine-tuning. Shading indicates 95\% confidence intervals across 6 random seeds.
} 
    \label{fig:bios}
\end{figure}

\begin{figure}[t]
    \centering
    \includegraphics[width=.87\columnwidth]{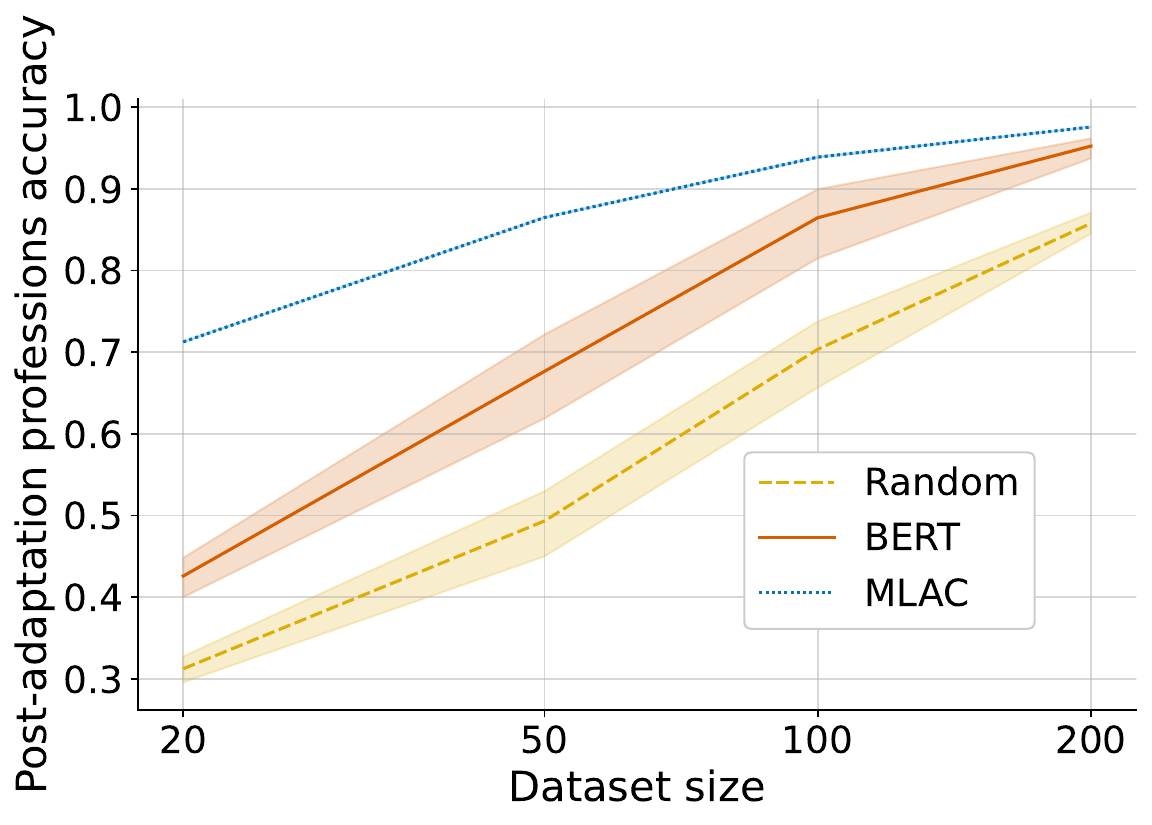}
    \caption{After fine-tuning the MLAC-blocked model on the desired task, few-shot performance exceeds both BERT and a randomly-initialized model. Note the MLAC objective includes training on the desired task, so this comparison clearly advantages MLAC; nonetheless, it provides evidence that there \textit{exists} a blocked initialization that can be effectively fine-tuned on the desired task. Discovering such an initialization without using desired task data in pre-training is an important direction for future work.}

        \label{fig:good}
\end{figure}

\subsection{Comparisons.} We compare \methodacc{} to the adversarial censoring (\textit{AC} in Fig.~\ref{fig:bios}) method from \citet{edwards2015censoring} as well as a model simply fine-tuned on the desired task (\textit{BERT (fine-tuned)} in Fig.~\ref{fig:bios}). For AC, an adversarial layer is learned on top of representation layers to predict the undesirable task. The gradient is then flipped to destroy undesirable information in the representation layer.
Notably, \methodacc{} with $K=0$ and with no calibration is equivalent to adversarial censoring.
We use a BERT-tiny model as our FM to save on compute costs~\citep{devlin2018bert,turc2020wellread} and use a linear classifier head for the tasks. Note that, as mentioned in Sec.~\ref{sec:method}, we focus on making sure that the professions task is unimpeded, so we directly train on cross-entropy loss as $\mathcal{L}_g$ during MLAC pre-training. For all models, the final achieved performance is retained for the desired professions task (see below and Figure~\ref{fig:good}). 

\subsection{Results.} 
\label{sec:results}
\begin{figure}[t]
    \centering
    \includegraphics[width=.87\columnwidth]{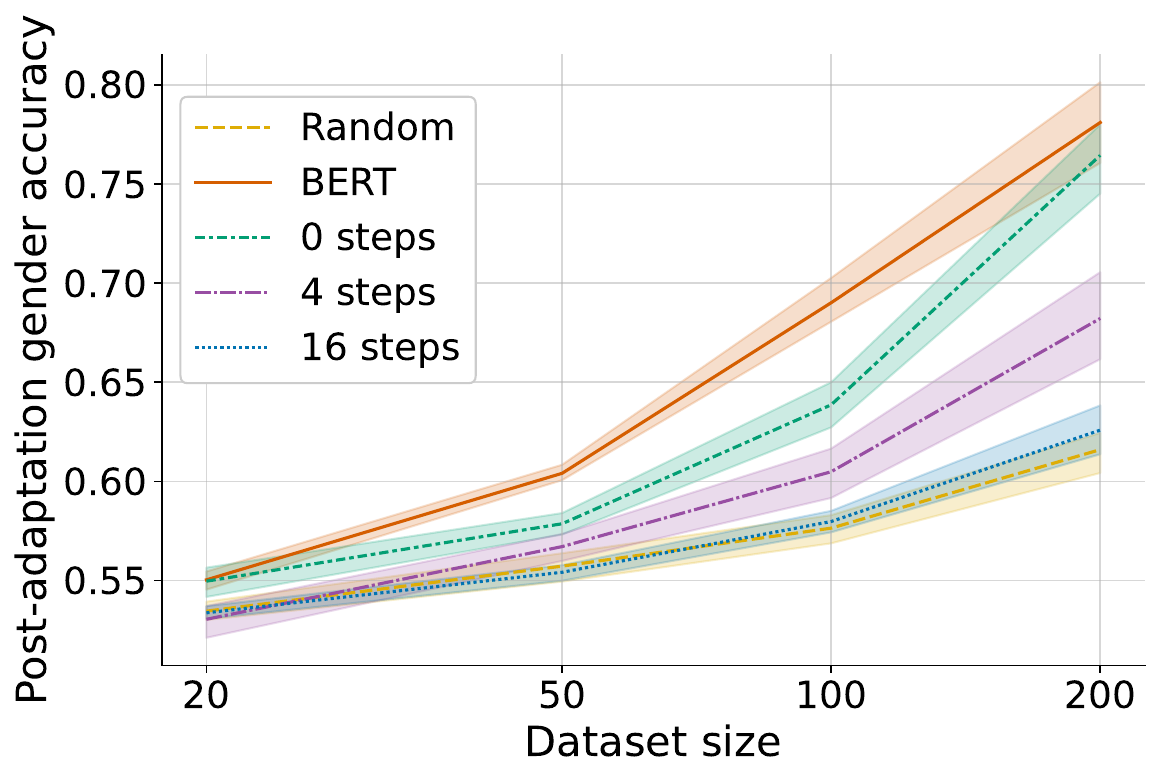}

    \caption{Evaluation of various inner loop depths during \methodacc{} training. Just 16 steps enables near-random performance, even though the adversary performs up to 1000 steps during fine-tuning.}

    \label{fig:steps}
\end{figure}

\begin{figure}[t]
    \centering
    \includegraphics[width=.87\columnwidth]{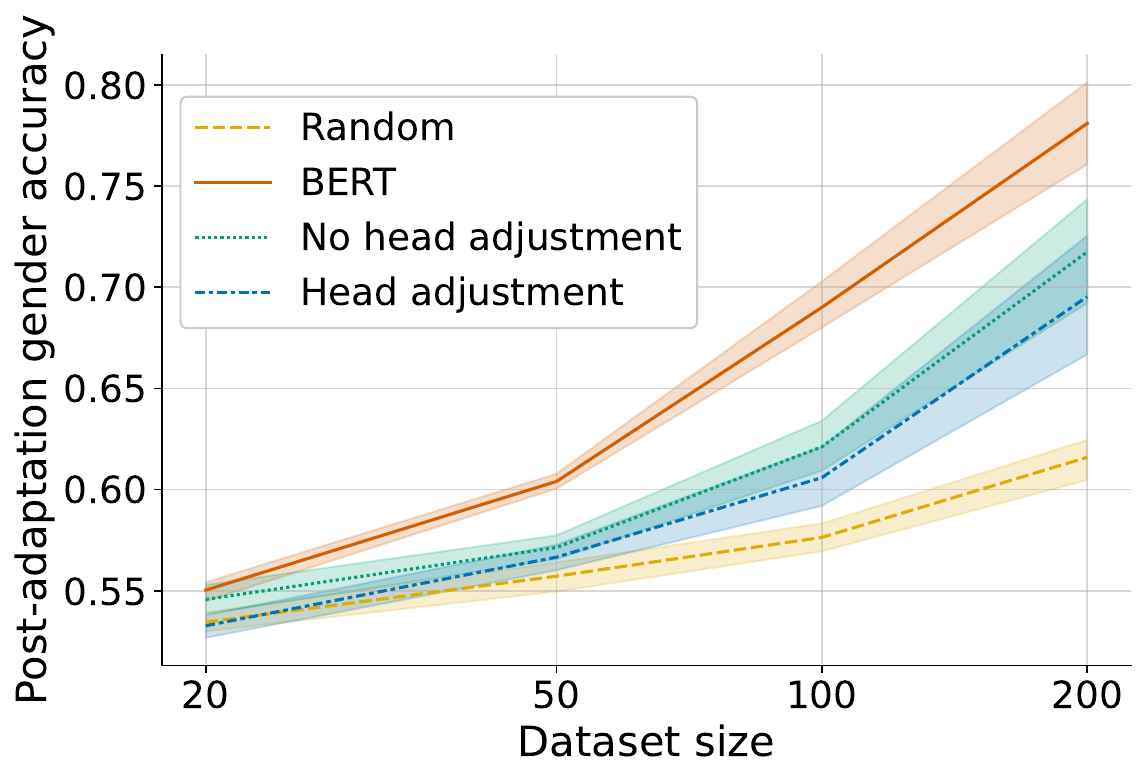}
    \caption{Ablating optimal adversary prediction calibration (or \textit{head adjustment}) during \methodacc{} training. Using optimally calibrated adversary predictions (modifying line 7 of Alg.~\ref{alg:mlac}) modestly improves blocking. Aggregated over 0, 4, and 16 steps.}

        \label{fig:calibration}
\end{figure}

Fig.~\ref{fig:bios} shows that MLAC returns nearly identical-to-random harmful task performance at all data regimes. Conversely, adversarial censoring (the equivalent of MLAC without calibration and $K=0$) does not appear to have any effect on post-fine-tuning harmful task performance. Fig.~\ref{fig:steps} shows the vital role played by the depth of the inner training loop of \methodacc{}, suggesting that \textit{a meta-learning process is genuinely necessary to impede harmful task performance}. To ensure that desired task performance is retained, we evaluate the blocked model on the desired task of profession classification, comparing with fine-tuning a pretrained BERT-tiny model and a random model. Fig.~\ref{fig:good} shows the result; \methodacc{} is clearly able to solve the task effectively, surpassing the few-shot performance of BERT-tiny.\footnote{Recall again that we use the desired task loss to counter-balance the task blocking mechanism, so this is expected. We do however use separate held-out subsets of data for final desired-task tuning and evaluation. As mentioned previously, our goal for the purposes of this initial exploration is to determine whether desired task performance can be retained while blocking a harmful task. Future work should examine generalization for retaining desired task adaptation performance across many desired tasks.} Finally, we find that head re-calibration may mildly improve blocking on average when pooled across all inner-loop step configurations (Fig.~\ref{fig:calibration}).

\section{Ethical Considerations and Limitations}

Before we conclude, we point out several other considerations and limitations.

First, while the goal of our approach is to make models safer overall, we recognize that value judgements will be made in deciding which tasks to block. Sometimes these judgement decisions can themselves encode biases and it requires an approach that takes into account a range of perspectives. Nonetheless, we argue that considering potential harmful dual-uses is an essential part of any modern model release process. Current standard licenses for foundation models already contain a list of restricted tasks~\citep{openrail2022,touvron2023llama}, but self-destructing models encode this directly into their optimization objective as well.

Second, it is necessary to collect data for harmful tasks to effectively block them. While this draws a direct parallel to security research, red-teaming, and white-hat hacking, there may be risks in aggregating this data. And there may be impacts on the well-being of potential annotators and security research members~\citep{mazeika2022would}. Sufficient precautions should be taken to mitigate these harms.

Third, there may be a risk of over-confidence in the self-destructing mechanism. While this paradigm adds a new tool to the safety toolkit, it does not completely prevent manipulation for every harmful task. And just like any other safety tool there will likely be a back-and-forth where adversaries learn to overcome some techniques. As such, self-destructing models can be combined with other safety mechanisms---structural or technical---to increase the costs of harmful dual-uses.

Fourth, our experiments demonstrate the functionality of self-destructing models in a constrained setting, but further work is needed to scale these approaches to more tasks, larger models, and more complicated settings. We believe this is an exciting new research direction, but requires more work to deploy at scale.

\section{Related Work}

A number of researchers have sought to address dual use risks by restricting points of control~\cite{flynn2020recommendations,brundage2018malicious,solaiman2019release,bommasani2021opportunities,shevlane2022structured,zwetsloot2019keeping}, despite there also being substantial benefits to open access~\cite{zhang2022opt,black2022gpt}. We aim to provide an alternative that allows for open access while still hindering bad actors.

Some work on AI safety has sought mechanisms to prevent agents from learning degenerate behaviors. \citet{orseau2016safely}, for example, seek to prevent a particular scenario where an agent learns to disable its off-switch so that it continues to collect reward. We on the other hand focus on preventing a different, broader, set of harmful behaviors: adaptation of pretrained models to harmful tasks.   

Closely related to our work are methods for de-biasing, editing, or removing harmful content from models. Like domain invariance approaches~\citep{ganin2015unsupervised,li2018domain,zhou2020domain,yao2022improving}, \citet{edwards2015censoring} use an adversarial approach to remove information from representations. 
\citet{ravfogel2022linear} and \citet{ravfogel2022adversarial} take a similar approach and find a projection on the final output layer of a pretrained model that removes gender-based biases from the model (and prevent recovery of those biases after that projection layer).
\citet{pryzant2018deconfounded} similarly use adversarial methods to remove confounds from representations.
Others have created model editing techniques to remove outdated or harmful content from pretrained models~\citep{Sinitsin2020Editable,Cao2021EditingFK,mitchell2022fast,mitchell2022memory}. 
While these other methods generally optimize for the information to be removed from the original model, we optimize for poor performance even \emph{after} adaptation of the original model to a harmful task. This can be accomplished via a meta-learning approach.

In the context of meta-learning, MAML \citep{finn2017model} and related algorithms \citep{li2017metasgd,lee2018gradient,park2019meta,zintgraf2018cavia,Flennerhag2020Meta-Learning} have shown that the desired \textit{post-}fine tuning behavior of a neural network can be effectively encoded in its \textit{pre}-fine tuning network initialization. While existing works have leveraged this ability in order to enable more rapid learning of new tasks, our work encodes a blocking mechanism into a network's initialization that \textit{prevents} effective adaptation on harmful tasks.

Finally, some scholars have tuned models to be safer by using reinforcement learning from human feedback and other approaches for incorporating human preferences, including \citet{bai2022training}, \citet{korbak2023pretraining}, \citet{ouyang2022training}, and others.

\section{Conclusion}

This work is only a first step in raising the cost for harmful dual uses of pretrained models through task blocking.
Future work might expand this study in at least four directions: \textit{scaling} the self-destructing model framework to larger FMs; studying \textit{generalization} of the learned blocking behavior to new (but related) datasets other than the one used during \methodacc{} meta-training; training/evaluating with \textit{stronger adversaries} that incorporate adaptation methods such as prefix tuning \citep{li2021prefixtuning}, adapter layers \citep{pmlr-v97-houlsby19a}, or others; and evaluating the preservation of desired task \textit{fine-tunability} for out-of-distribution tasks. Future work might also introduce concealed architectural changes that hide self-destruct triggers in the network but are more robust to adversarial mechanisms. We hope self-destructing models can become one tool enabling model developers to share their artifacts while minimizing dual use risks.

\begin{acks}
We thank Rishi Bommasani, Siddharth Karamcheti, and Jieru Hu for helpful discussion and feedback. PH is supported by an Open Philanthropy AI Fellowship. EM is supported by a Knight-Hennessy Graduate Fellowship. CF and CM are CIFAR Fellows.
\end{acks}

\bibliographystyle{ACM-Reference-Format}
\bibliography{sample-base}

\end{document}